\documentclass[keeplastbox, letterpaper, 10 pt, conference]{ieeeconf}  

\IEEEoverridecommandlockouts                              

\makeatletter
\let\NAT@parse\undefined
\makeatother                                 

\usepackage{graphics} 
\usepackage{graphicx}
\usepackage{bm} 
\usepackage{amsmath} 
\usepackage{amssymb}  
\usepackage{flushend}
\usepackage{float}
\usepackage{subfig}
\usepackage{makecell}
\usepackage[normalem]{ulem}
\usepackage{hyperref}
\usepackage{color}

\usepackage{amsfonts}
\usepackage{algorithm,algpseudocode}

\usepackage{gensymb}
\usepackage[tbtags]{mathtools}

\usepackage[pdftex,dvipsnames]{xcolor}

\usepackage{todonotes}
\usepackage{units}

\usepackage{tikz}
\usepackage{textcomp}

\usepackage[comma,numbers,compress]{natbib}

\usepackage[firstpage=true]{background}

\newcommand{\norm}[1]{\left\lVert#1\right\rVert}

\SetBgContents{\textbf{Accepted final version.} In \textit{2018 IEEE International Conference on Robotics and Automation (ICRA)}}
\SetBgScale{1}
\SetBgColor{black}
\SetBgAngle{0}
\SetBgOpacity{1}
\SetBgPosition{current page.south}
\SetBgVshift{1cm}

\title{\LARGE \bf
Transferring Grasping Skills to Novel Instances\\ by Latent Space Non-Rigid Registration}

\author{Diego Rodriguez, Corbin Cogswell, Seongyong Koo, and Sven Behnke
\thanks{$^{1}$All authors are with the Autonomous Intelligent Systems (AIS) Group, Computer Science Institute VI, University of Bonn, Germany,
        {\tt\small \{rodriguez, koo, behnke\}@ais.uni-bonn.de}}%
}

\begin{document}

\maketitle
\thispagestyle{empty}
\pagestyle{empty}

\begin{abstract}
Robots acting in open environments need to be able to handle novel objects.
Based on the observation that objects within a category are often similar in their shapes and usage, we propose an approach for transferring grasping skills from known instances to novel instances of an object category.
Correspondences between the instances are established by means of a non-rigid registration method that combines the Coherent Point Drift approach with subspace methods. 

The known object instances are modeled using a canonical shape and a transformation which deforms it to match the instance shape.
The principle axes of variation of these deformations define a low-dimensional latent space.
New instances can be generated through interpolation and extrapolation in this shape space.
For inferring the shape parameters of an unknown instance, an energy function expressed in terms of the latent variables is minimized.
Due to the class-level knowledge of the object, our method is able to complete novel shapes from partial views.  
Control poses for generating grasping motions are transferred efficiently to novel instances by the estimated non-rigid transformation.  
\end{abstract}

\section{Introduction}
\label{sec:introduction}

\begin{figure}
    \centering
    \includegraphics[width=1.0\linewidth]{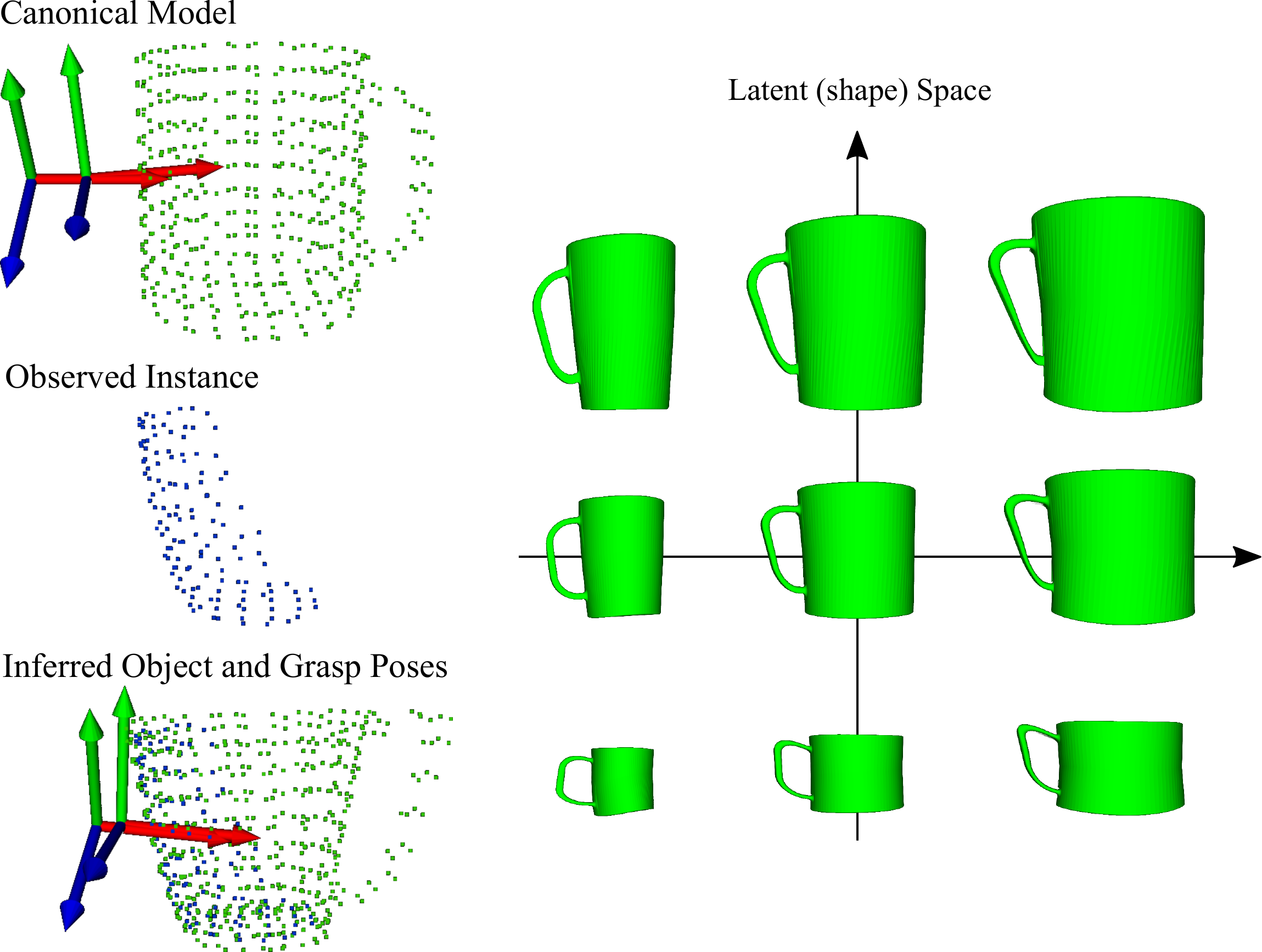}
    \caption{\footnotesize Control poses are transferred to the shape of a novel instance by latent space non-rigid registration and used to generate a valid grasping motion.}
    \label{fig:transfer}
\end{figure}

People can be given a screwdriver that they never saw before, and they will immediately know how to grasp and operate it by transferring previous manipulation knowledge to the novel instance. 
While this transfer happens effortless in humans, achieving such generalization in robots is challenging. 
Manipulating novel instances of a known object category is still an open problem. 
Although the manipulation of known objects can be planned offline, many open-world applications require the manipulation of unknown instances. 
Our approach transfers manipulations skills to novel instances by means of a novel latent space non-rigid registration (Fig. \ref{fig:transfer}).

Robots are often equipped with RGB-D sensors to perceive their environment in 3D.
In order to reconstruct the full shape of an object---desirable for planning grasping---multiple views of the object must be taken and fused into a single 3D model.
However, robots are not always able to obtain the required views for generating the full model, because of obstructions or unreachable observation poses, e.g., an object lying at the inner corner of a shelf. 
Fortunately, it is frequently not necessary to measure occluded object parts if they can be inferred from prior object knowledge. 

In this paper, we propose a method for generating grasping motions for novel instances from a single view by making use of category-level extrinsic shape information residing in a learned latent space.
Our method accumulates object shape from multiple known instances of a category in a canonical model. 
The learned latent space of shape variations enables a category-specific shape-aware non-rigid registration procedure that establishes correspondences between the novel view and the canonical model. 
By linearly interpolating between low-dimensional shape parameters of the known instances, and by extrapolation in this shape space, a manifold of plausible shapes belonging to the modeled category can be generated.

Our method finds a transformation from the canonical model to a view of a novel instance in the latent space---linearly interpolated and extrapolated from other transformations found within the class---which best matches the observed 3D points.
This estimates the shape parameters of the novel instance and allows for inference of its occluded parts.
The non-rigid transformation maps control poses from the canonical model to the novel instance, which are then used to generate the grasping motion for it.

The remainder of this paper is organized as follows. In the next section, we discuss related works. Section~\ref{sec:statement} gives an overview of our approach. The necessary background is then presented in Section~\ref{sec:cpd}. In Section~\ref{sec:method}, we describe our approach in detail. 
We evaluated our method in multiple experiments which are presented in Section~\ref{sec:experiments}.

\section{Related Work}
\label{sec:related_work}

\subsection{Non-Rigid Registration}
\label{sec:Registration}

Since \citet{chen1992object} introduced the Iterative Closest Point (ICP) algorithm, numerous variations have been proposed for rigid registration~\cite{rusinkiewicz2001efficient, holz2015registration}.
For non-rigid registration, priori restrictions or regularization on the motion or deformation of the points between sets are often imposed. Different transformation priors such as isometry~\cite{bronstein2006efficient, tevs2009isometric, ovsjanikov2010one}, elasticity~\cite{haehnel2003extension}, conformal maps~\cite{levy2002least, zeng2010dense, kim2011blended}, thin-plate splines~\cite{allen2003space, brown2007global}, and Motion Coherence Theory~\cite{myronenko2010point} have been used to allow for or to penalize different types of transformations. 

Many methods use non-rigid registration for surface reconstruction~\citep{li2008global, li2009robust, sussmuth2008reconstructing, wand2007reconstruction, wand2009efficient, newcombe2015dynamicfusion}.
\citet{brown2007global, li2008global} proposed surface reconstruction methods for global non-rigid registration.
Methods such as \citet{li2009robust, zollhofer2014real} use a Kinect depth camera to capture an initially low-resolution 3D surface and use non-rigid registration to continuously add higher-frequency details to the model with each new frame.

Recently, \citet{newcombe2015dynamicfusion} proposed a non-rigid dense surface reconstruction method using dense non-rigid registration. 
From a sequence of depth images, they calculate a dense 6-dimensional warp field between frames. 
They then undo the transformations between each frame and rigidly fuse the scans into one canonical shape. 
The dense warp field is estimated by minimizing an energy function.
Although this method is able to deform a scene in real-time toward a canonical shape, 
the use optical flow constraints makes this method inadequate for large deformations or strong changes in illumination and color.

\subsection{Class-Level Shape Spaces}
\label{sec:Class-Level_Shape_Spaces}
To create a parameterized space of shapes, several methods have been proposed. 
\citet{blanz1999morphable} create a morphable model of faces able to create novel faces and to interpolate between faces using a few parameters.
Similarly, \citet{allen2003space} create a shape space of human bodies using human body range scans with sparse 3D markers.
\citet{hasler2009statistical} extend this space to include pose, creating a unified space of both pose and body shape. This allows them to model the surface of a body in various articulated poses more accurately.

Other approaches as the one presented by \citet{nguyen2011optimization} establish shape correspondences by creating collections of similar shapes and optimizing the mapping at a global scale.
\citet{huang2012optimization} also use collections of shapes to enforce global consistency. They create a small collection of base shapes from which correspondences are established between all other shapes.

\citet{burghard2013compact} propose an approach to estimate dense correspondences on a shape given initial coarse correspondences. They use the idea of minimum description length to create a compact shape space of related shapes with strongly varying geometry. 
However, this method does not perform well in the presence of noise or incomplete scans.
\citet{engelmann2016joint} learn a compact shape manifold which represents intra-class shape variance, allowing them to infer shape in occluded regions or regions where data might be missing or noisy such as on textureless, reflective, or transparent surfaces. 
This approach however does not give correspondences between points and do not offer any kind of transformation, 
which limits its applicability to transferring grasping knowledge.

\subsection{Transferring Grasping Skills}
\citet{stuckler2011real} proposed a method for manipulation skill transfer using non-rigid registration. 
The registration finds a non-rigid transformation between a known instance where manipulation information---such as grasp poses or motion trajectories---are available and a novel instance from the same class. 
The non-rigid transformation is applied to map trajectory control poses towards the newly observed instance, which allows for use of the novel object. 
We extend this work by modeling shape and grasping not for single known instances, but within a category, which allows for learning typical variations of a canonical model.

Alternative methods as in \cite{vahrenkamp} transfer grasp poses by segmenting the objects in primitive shapes according to their RGB-D appearance. 
Because this method is based on templates primitives, it is not able to handle occlusions efficiently.
In \cite{stouraitis} and in \cite{amor}, grasp poses are transferred using the same contact warp method which minimizes the distance between assigned correspondences. 
However, both approaches require a fully observed object and focus only on the final result of grasping, namely on grasp poses, while our approach addresses the process of grasping.

\section{Problem Statement}
\label{sec:statement}
We propose a new approach for transferring grasping skills based on a novel method for non-rigid registration.
This method incorporates category-level information by learning a shape space of the category. 
By incorporating this information into the registration, we can avoid unlikely shapes and focus on deformations actually observed within the class. 
Thus, the resulting grasping motion is able to tolerate the noise and occlusions typical of data measured by 3D sensors. 
Furthermore, the generated space can be used to create novel instances and to interpolate and extrapolate between previous ones. 

\begin{figure*}[t]
	\centering
	\includegraphics[width=1.0\linewidth]{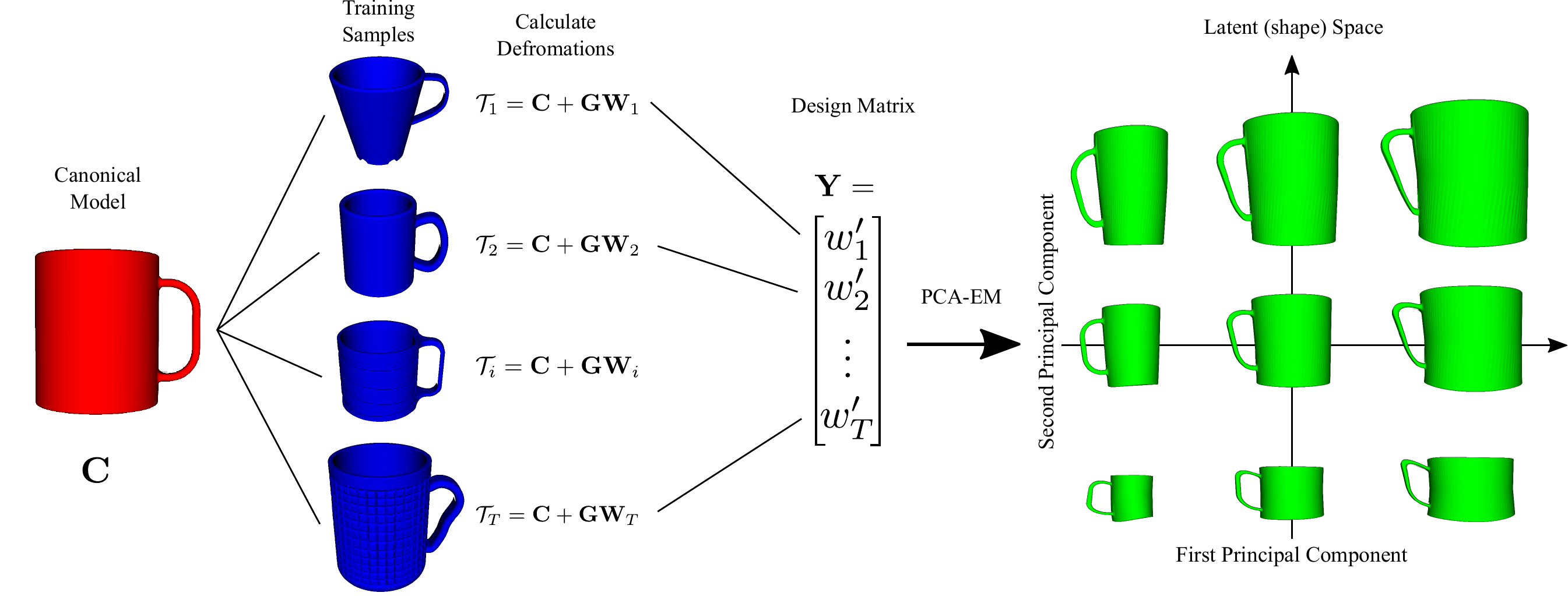} 
	\caption{\footnotesize Learning of the latent space. First, the deformations $\mathbf{W}_i$ between each instance $\mathcal{T}_i$ and the canonical model $\mathbf{C}$ are calculated using CPD. These deformations are then expressed as column vectors $\mathbf{w}_i$ and assembled into the design matrix $\mathbf{Y}$. Using PCA-EM, the principal components which constitute the latent shape space are extracted.}
	\label{fig:overview}
\end{figure*}

\begin{figure}[t]
    \centering
    \includegraphics[width=0.9\linewidth]{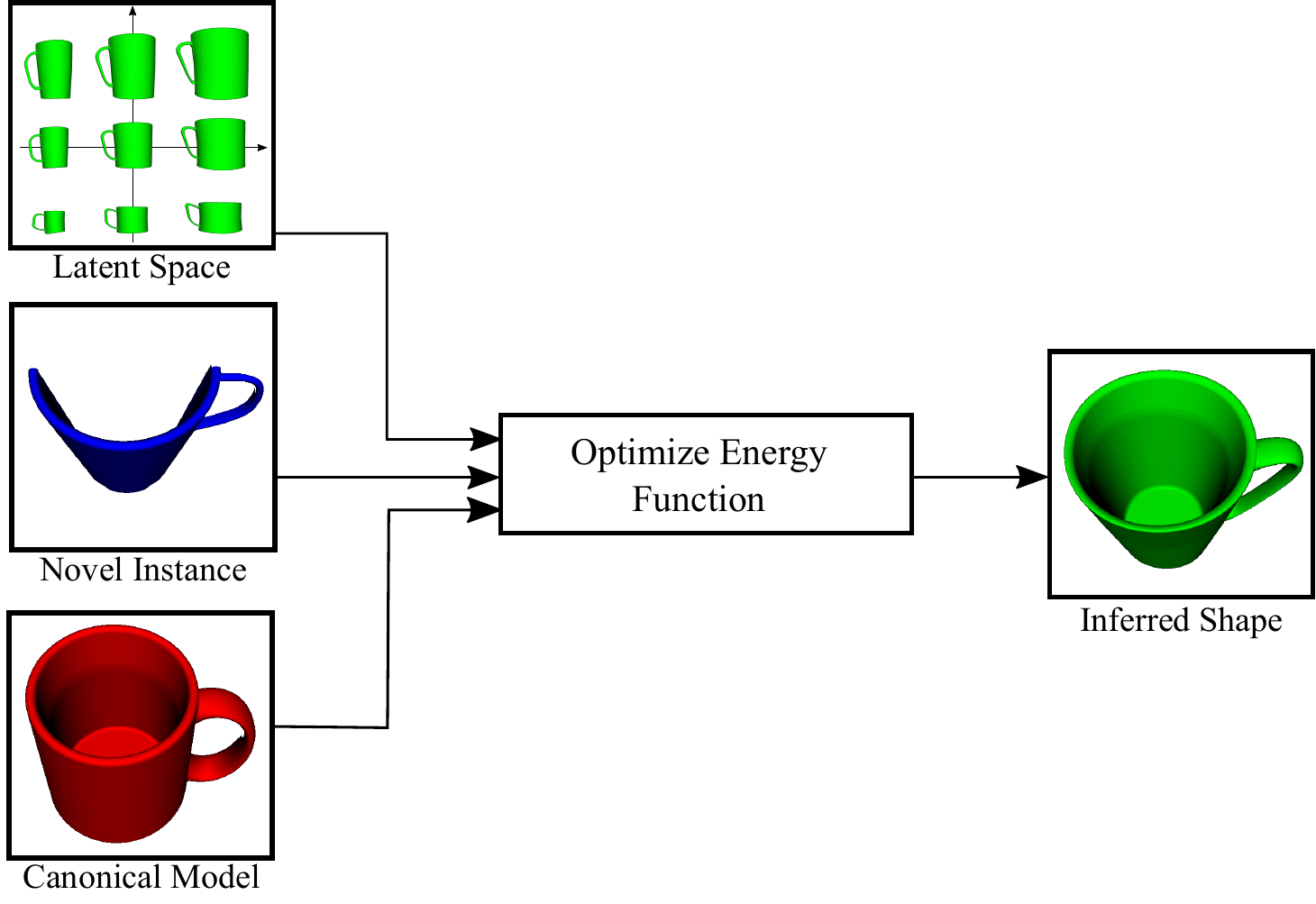}
    \caption{\footnotesize The canonical shape (red) is matched against a partially-occluded target shape (blue) by optimizing Eq. (\ref{eq:Energy}) in terms of the latent parameters plus a rigid transformation. The resulting shape is shown in green on the right.}
    \label{fig:inference}
\end{figure}

Our method is divided into two phases: a learning phase and an inference phase (Fig. \ref{fig:overview} and \ref{fig:inference}). 
In the learning phase, the objective is to create a class-specific linear model of the transformations that a class of objects can undergo. 
We do this by first selecting a single model to be a canonical instance of the class, and then we find the transformations relating this instance to all other instances of the class using Coherent Point Drift (CPD) \citep{myronenko2010point}. 
We then find a linear subspace of these transformations, which becomes our transformation model for the class. 
In the inference phase, the objective is: given a newly observed instance, search this subspace of transformations to find the transformation which best relates the canonical shape to the observed instance. 
Grasping information from the canonical model is also transformed to the observed instance and used to generate the grasping motion of the robot.

The transformations are represented by a dense deformation field plus a single rigid transformation; the latter is included to account for small global misalignments. 
This allows points defined in the space of the canonical shape to be transformed into the space of an observed instance. 
These points do not need to be known a priori for the learning phase and can be added at any time after the registration has completed. 

The main contributions of this paper are:
\begin{enumerate}
    \item a novel learned non-rigid registration method that finds a low-dimensional space of deformation fields; this method can be used to interpolate and extrapolate between instances of a class of objects to create novel instances and
    \item the application of this non-rigid registration method to transfer grasping skills.
\end{enumerate}

\section{Coherent Point Drift}
\label{sec:cpd}
We briefly review the Coherent Point Drift (CPD) \citep{myronenko2010point} method on which our non-rigid registration is based.

Given two point sets, a template set $\mathbf{S}^{[t]} = (\mathbf{s}^{[t]}_1, ..., \mathbf{s}^{[t]}_M)^T$ and a reference point set $\mathbf{S}^{[r]} = (\mathbf{s}^{[r]}_1, ..., \mathbf{s}^{[r]}_N)^T$, CPD tries to estimate a deformation field mapping the points in $\mathbf{S}^{[t]}$ to $\mathbf{S}^{[r]}$.
The points in $\mathbf{S}^{[t]}$ are considered centroids of a Gaussian Mixture Model (GMM) from which the points in $\mathbf{S}^{[r]}$ are drawn. 
CPD seeks to maximize the likelihood of the GMM while imposing limitations on the motion of the centroids. 
CPD imposes a smoothness constraint on the deformation of the points in the form of motion coherence, which is based on Motion Coherence Theory \citep{yuille1988motion}. 
The idea is that points near each other should move coherently and have a similar motion to their neighbors.
Maximizing the likelihood of the GMM is equivalent to minimizing the energy function:
\begin{equation}
    \label{eq:Deformation_Energy}
    E(\mathbf{S}^{[t]}, \bm{\psi}) = -\sum^N_{n=1}{\log{\sum^M_{m=1}{e^{-\frac{1}{2\sigma^2} \norm{\mathbf{s}^{[r]}_n-\mathcal{T}(\mathbf{s}^{[t]}_m, \bm{\psi})}^2}}}} + \frac{\lambda}{2}\phi(\mathbf{S}^{[t]})
\end{equation}
where $\mathcal{T}(\mathbf{s}^{[t]}_m, \bm{\psi})$ is a parametrized transformation from the template point set to the reference set. 
The first term of Eq. (\ref{eq:Deformation_Energy}) penalizes the distance between points after applying the transformation $\mathcal{T}$, and the second term is a regularization term which enforces motion coherence.

For the non-rigid case, the transformation $\mathcal{T}$ is defined as the initial position plus a displacement function $v$: 
\begin{equation}
    \label{eq:Deformation_Vector}
    \mathcal{T}(\mathbf{S}^{[t]},v) = \mathbf{S}^{[t]} + v(\mathbf{S}^{[t]}),
\end{equation}
where $v$ is defined for any set of $D$-dimensional points $\mathbf{Z}_{N\times D}$ as:
\begin{equation}
    \label{eq:Deformation_Field}
    v(\mathbf{Z}) = G(\mathbf{S^{[t]}}, \mathbf{Z})\mathbf{W}.
\end{equation}
$G(\mathbf{S^{[t]}}, \mathbf{Z})$ is a Gaussian kernel matrix which is defined element-wise as:
\begin{equation}
    \label{eq:Gaussian_Kernel}
	g_{ij} = G(\mathbf{s}^{[t]}_i, \mathbf{z}_j) = \exp^{ -\frac{1}{2\beta^2} \norm{ \mathbf{s}^{[t]}_i-\mathbf{z}_j} ^2},
\end{equation}
$\mathbf{W}_{M\times D}$ is a matrix of kernel weights, 
and $\beta$ is a parameter that controls the strength of interaction between points. 
An additional interpretation of $\mathbf{W}$ is as a set of $D$-dimensional deformation vectors, each associated with one of the $M$ points of $\mathbf{S}^{[t]}$.
For convenience in the notation, $\mathbf{G}_{M\times M}$ will be denoted $\mathbf{G}(\mathbf{S}^{[t]},\mathbf{S}^{[t]})$.
Note that $G(\cdot,\cdot)$ can simply be computed by Eq. (\ref{eq:Gaussian_Kernel}), but the matrix $\mathbf{W}$ needs to be estimated. 

CPD uses an Expectation Maximization (EM) algorithm derived from the one used in Gaussian Mixture Models \citep{bishop1995neural} to minimize Eq. (\ref{eq:Deformation_Energy}).
In the E-step, the posterior probabilities matrix $\mathbf{P}$ is estimated using the previous parameter values. To add robustness to outliers, an additional uniform probability distribution is added to the mixture model. This matrix $\mathbf{P}$ is defined element-wise as:
\begin{equation}
    \label{eq:E-Step}
    p_{mn} = \frac{e^{-\frac{1}{2\sigma^2}\norm{\mathbf{s}^{[r]}_n-(\mathbf{s}^{[t]}_m+G(m,\cdot)\mathbf{W})}^2}} {\sum^M_{m=1}{e^{-\frac{1}{2\sigma^2}\norm{\mathbf{s}^{[r]}_n-(\mathbf{s}^{[t]}_m+G(k,\cdot)\mathbf{W})}^2}}+\frac{\omega}{1-\omega}\frac{(2\pi\sigma^2)^{\frac{D}{2}}}{N}}
\end{equation}
where $\omega$ reflects the assumption on the amount of noise.

In the M-step, the matrix $\mathbf{W}$ is estimated by solving the equation:
\begin{equation}
    \label{eq:M-Step}
    (\mathbf{G}+\lambda\sigma^2d(\mathbf{P1})^{-1})\mathbf{W} = d(\mathbf{P1})^{-1}\mathbf{P}\mathbf{S}^{[r]} - \mathbf{S}^{[t]}
\end{equation}
where $\mathbf{1}$ represents a column vector of ones and $d(\cdot)^{-1}$ is the inverse diagonal matrix.

In our method, we use the canonical shape $\mathbf{C}$ for the deforming template shape $\mathbf{S}^{[t]}$ and each training example $\mathbf{T}_i$ as the reference point set $\mathbf{S}^{[r]}$. Hence, the transformations $\mathcal{T}_i$ are defined as 
\begin{equation}
\label{eq:defomation_field}
\mathcal{T}_i(\mathbf{C},\mathbf{W}_i) = \mathbf{C} + \mathbf{G}\mathbf{W}_i
\end{equation}

where $\mathbf{W}_i$ is the $\mathbf{W}$ matrix computed by taking training example $\mathbf{T}_i$ as the reference point set $\mathbf{S}^{[r]}$.
\section{Method}
\label{sec:method}
Our learning-based approach has an initial training phase in which the latent space for a class of objects is built. 
With the latent space of the object class, inference can be performed to find a transformation which relates one of the shapes from the class called the canonical shape to fully or partially-observed novel instances. The transformation is represented by a dense deformation field plus a single rigid transformation. Finally, we transform all the grasping information to the new instance.

\subsection{Categories and Shape Representation}
\label{sec:Classes_and_Shape_Representation}
\begin{figure*}
    \centering
    \includegraphics[width=1.0\linewidth]{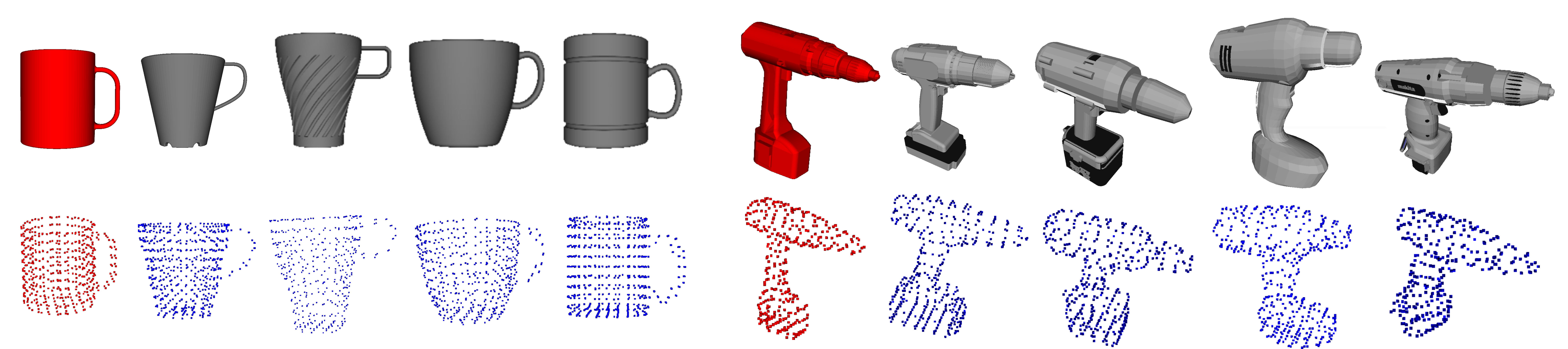} 
    \caption{\footnotesize Original meshes and the uniformed sampled point clouds of instances of the \textit{Mug} and \textit{Drill} categories. The canonical models are colored red.}
    \label{fig:Class_Definition}
\end{figure*}

We define a category or class as a set of objects which share the same topology and a similar extrinsic shape. 
A category is composed of a set of training sample shapes and a canonical shape $\mathbf{C}$ that will be deformed to fit the shape of the training and observed samples.
To represent a shape, we use point clouds, which can be generated from meshes by ray-casting from several viewpoints on a tessellated sphere and then down-sampling with a voxel grid filter. 
The canonical shape can be chosen by heuristics or chosen as the shape with the lowest reconstruction energy after finding the deformation field of all other training examples.
Fig. \ref{fig:Class_Definition} shows both the mesh and the resulting uniformly-sampled point clouds of object samples of two different categories. 

Each class must specify a canonical pose and reference frame, which are essential for initial alignment. 
For example: for \textit{Mugs}, we can align all instances such that their top is open upward with all the handles pointing towards the right. 

\subsection{Low-Dimensional Deformation Field Manifold}
Once we have a set of training examples and a canonical shape, we need to find the deformations from the canonical shape to all other training shapes. Here we make use of the Coherent Point Drift method \citep{myronenko2010point}.

CPD provides a dense deformation field, allowing us to find deformation vectors for novel points, even those added after the field is created, which in turn allows us to apply the method for transferring grasping skills. 
Additionally, CPD allows us to create a feature vector representing the deformation field. 
Importantly, this vector has the same length for all training examples and elements in the vector correspond with the same elements in another. 
This will allow us to construct a latent space in later steps.
 
From Eq. (\ref{eq:defomation_field}), we see that the deformation field function is defined by the matrices $\mathbf{G}$ and $\mathbf{W}$. 
Moreover, from Eq. (\ref{eq:Gaussian_Kernel}) we see that $\mathbf{G}$ is a function only of the canonical shape and remains constant for all training examples. 
Thus, the entire uniqueness of the deformation field for each training example is captured by the matrix $\mathbf{W}$.

For each training example $\mathbf{T}_i$, we take the matrix $\mathbf{W}_i$ from its deformation field and convert it into a row vector $\mathbf{y}_i\in \mathbb{R}^{p=M\times D}$, which becomes the feature descriptor of that deformation field. 
The vectors are then normalized resulting in zero-mean and unit-variance vectors which are then assembled into a design matrix $\mathbf{Y}$. 
Finally, we apply Principle Component Analysis (PCA) on this matrix to find a lower-dimensional manifold of deformation fields for this class. 

PCA finds a matrix $\mathbf{L}_{p \times q}$ of principle components that can be used to estimate a matrix of $n$ observation vectors $\mathbf{\hat{Y}}_{n\times p}$ given a small set of $q$ latent variables, i.e., $q <\!\!< p$:
\begin{equation}
    \label{eq:To_Observed}
    \mathbf{\hat{Y}} = \mathbf{X}\mathbf{L}^T. 
\end{equation}

For a new normalized set of observations $\mathbf{Y_o}$, the latent variables can be found by:
\begin{equation}
    \label{eq:To_Latent}
    \mathbf{X} = \mathbf{Y_o}\mathbf{L}.
\end{equation}
We find the matrix $L$ using the PCA Expectation Maximization (PCA-EM) algorithm \citep{roweis1998algorithms}. We choose this method over analytical algorithms due to its superior performance in situations with high dimensions and scarce data and also because of its greater numerical stability.

Much like with CPD, we alternate between an E- and M-step. The E-step is given by:
\begin{equation}
    \label{PCA_E-Step}
    \mathbf{X} = \mathbf{Y}\mathbf{L}^T(\mathbf{L}\mathbf{L}^T)^{-1}
\end{equation}
whereas the M-step is defined by:
\begin{equation}
    \label{PCA_M-Step}
    \mathbf{L} = (\mathbf{X}^T\mathbf{X})^{-1}\mathbf{X}^T\mathbf{Y}.
\end{equation}

This method is shown to converge to a local minimum using standard EM convergence proofs \citep{dempster1977maximum}. Additionally, it has been shown that the only stable local extremum is the global maximum \citep{tipping1999mixtures, tipping1999probabilistic}, meaning the algorithm will always converge to the correct result with enough iterations.

Using Eq. (\ref{eq:To_Latent}), a deformation field can now be described by only $q$ latent parameters.
Similarly, any point $\mathbf{x}$ in the latent space can be converted into a deformation field transformation by first applying Eq. (\ref{eq:To_Observed}) and converting the result into a $\mathbf{W}_{M \times D}$ matrix after the respective denormalization. 
Thus, moving through the $q$-dimensional space linearly interpolates between the deformation fields.

The matrix $\mathbf{L}$ and the canonical shape $\mathbf{C}$ together represent the transformation model for a class. 
Figure \ref{fig:overview} gives an overview of the training phase which is also summarized in Algorithm \ref{alg:Training}.
\begin{algorithm}[b]
    \caption{Building the latent space for a Category}
    \label{alg:Training}
        \textbf{Input:} A set of training shapes $\mathbf{E}$ in their canonical pose and reference frame
        \begin{algorithmic}[1]
            \State Select a canonical shape $\mathbf{C}$ via heuristic or pick the one with the lower reconstruction energy.
            \State Estimate the deformation fields between the canonical shape and the other training examples using CPD.
            \State Concatenate the resulting set of $\mathbf{W}$ matrices from the deformation fields into a design matrix $\mathbf{Y}$.
            \State Perform PCA on the design matrix $\mathbf{Y}$ to compute the latent space of deformation fields.
        \end{algorithmic}
        \textbf{Output:} A canonical shape $\mathbf{C}$ and a latent space of deformation fields represented by $\mathbf{L}$.
\end{algorithm}

\subsection{Shape Inference for Novel Instances}
\label{sec:Inference}
With the transformation model, we can now start registering the canonical shape to novel instances in order to estimate the underlying transformation. 
The parameters of the transformation are given by the $q$ parameters of the latent vector $\mathbf{x}$ plus an additional seven parameters of a rigid transformation $\bm{\theta}$. 
The rigid transformation is meant to account for minor misalignments in position and rotation between the target shape and the canonical shape at the global level. 

We concurrently optimize for shape and pose using gradient descent. 
We expect local minima, especially with regard to pose, therefore our method, as CPD and ICP, requires an initial coarse alignment of the observed shape. 
We want to find an aligned dense deformation field which when applied to the canonical shape $\mathbf{C}$, minimizes the distance between corresponding points in the observed shape $\mathbf{O}$. Specifically, we want to minimize the energy function:
\begin{equation}
    \label{eq:Energy}
    E(\mathbf{x},\bm{\theta}) = -\sum^{M}_{m=1}{\log{\sum^{N}_{n=1}{e^{\frac{1}{2\sigma^2}\norm{\mathbf{O}_n-\Theta(\mathcal{T}_m(\mathbf{C}_m,\mathbf{W}_m(\mathbf{x})),\bm{\theta})}^2}}}},
\end{equation}
where the function $\Theta$ applies the rigid transformation given $\bm{\theta}$ parameters.

When the minimum is found, we can transform any point or set of points into the observed space by applying the deformation field using Eq. (\ref{eq:Deformation_Field}) and Eq. (\ref{eq:Deformation_Vector}) and then applying the rigid transformation $\Theta$. 
Algorithm \ref{alg:Inference} summarizes the inference process. 

\subsection{Transferring Grasping Skills}
From experiences with an anthropomorphic robotic hand, namely the multi-fingered Schunk hand \citep{ruehl}, we observed that a functional grasping cannot be described only by the final joint configuration of the entire system (manipulator plus hand). 
During the motion, the hand can make use of geometrical and frictional constraints to increase the chances of a successful grasping. 
For instance, for grasping drills, we can first make a lateral contact with the palm followed by a contact of the thumb with the rear part of tool, creating static frictional forces that may help during the grasp. 
Therefore, we use the term \textit{grasping} when we refer to the complete action and the term \textit{grasp} for the result of this action.

We represent a grasping action as a set of parametrized motion primitives.
The parameters of the motion primitives are poses expressed in the same coordinate system of the shape of the object.
These poses need to be defined only for the canonical model.
For new instances, the poses are found by warping the poses of the canonical model to the instance.
Because the warping process can violate the orthogonality of the orientation, we orthonormalize the warped quaternions.
Additional parameters of the motion primitives such as velocities and accelerations can also be derived from the warped poses.

\begin{algorithm}[t]
	\caption{Shape Inference for a Novel Instance}
	\label{alg:Inference}
	\textbf{Input:} Transformation model ($\mathbf{C}$, $\mathbf{L}$) and observed shape $\mathbf{O}$
	\begin{algorithmic}[1]
		\State Compute matrix $\mathbf{G} = G(\mathbf{C}, \mathbf{C})$ (Eq. \ref{eq:Gaussian_Kernel}).
		\State Use gradient descent to estimate the parameters of the underlying transformation ($\mathbf{x}$ and $\bm{\theta}$) until the termination criteria is met. To calculate the value of the energy function, in each iteration: 
		
		-  Using the current values of $\mathbf{x}$ and $\bm{\theta}$:
		\begin{enumerate}
			\item Use Eq. (\ref{eq:To_Observed}) to create vector $\mathbf{\hat{Y}}$ and convert it into matrix $\mathbf{W}$.
			\item Use Eq. (\ref{eq:Deformation_Field}) and Eq. (\ref{eq:Deformation_Vector}) to deform $\mathbf{C}$.
			\item Apply the rigid transformation $\Theta$ to the deformed $\mathbf{C}$.
		\end{enumerate}
	\end{algorithmic}
	\textbf{Output:} Non-rigid transformation given by deformation field description $\mathbf{W}$ and rigid transform $\bm{\theta}$.
\end{algorithm}

\section{Experimental Results}
\label{sec:experiments}
We tested our method on two categories: \textit{Mugs} and \textit{Drills}. 
We obtained the models from two online CAD databases: GrabCad\footnote{\url{https://grabcad.com/library}} and the 3DWarehouse\footnote{\url{https://3dwarehouse.sketchup.com/}}. 
The meshes were converted into point clouds by ray-casting from several viewpoints on a tessellated sphere and down-sampling with a voxel grid filter.
We trained the \textit{Mug} and \textit{Drill} models with their canonical shape and additional 21 and 9 samples, respectively.
The meshes of some of the samples together with the uniformly sampled point clouds are shown in Fig.~\ref{fig:Class_Definition}.

We evaluated the robustness of our method to noise, misalignment, and occlusion and compared our results against results given by CPD. 
Noise was added to each point by randomly sampling a point from a normal distribution and scaling it by a noise factor.
Misalignment was generated by adding a rigid transformation to the observed shape. 
For the translation, we uniformly sampled a three-dimensional unit vector and multiplied it by the following factors: $[0.01, 0.02, 0.03, 0.04, 0.05]$.
For the orientation, a three-dimensional unit rotation axis was uniformly sampled and combined with the following angles: $[\pi/4, \pi/8, 3\pi/16, \pi/4, 3\pi/8$], making use of the axis-angle representation. 
Each test was run on a full view of the object and on six different partial views of the object. 
To obtain the partial views, we used ray-casting on a single view of a tessellated sphere. 

In all experiments, we parametrized CPD with the following values $\beta = 1$ and $\lambda = 3$. 
For creating the shape space, the number of latent variables was set to capture at least 95\% of the variance of each class, and the class canonical shapes were selected by experts.

\begin{figure}
    \centering
    \includegraphics[width=1.0\linewidth]{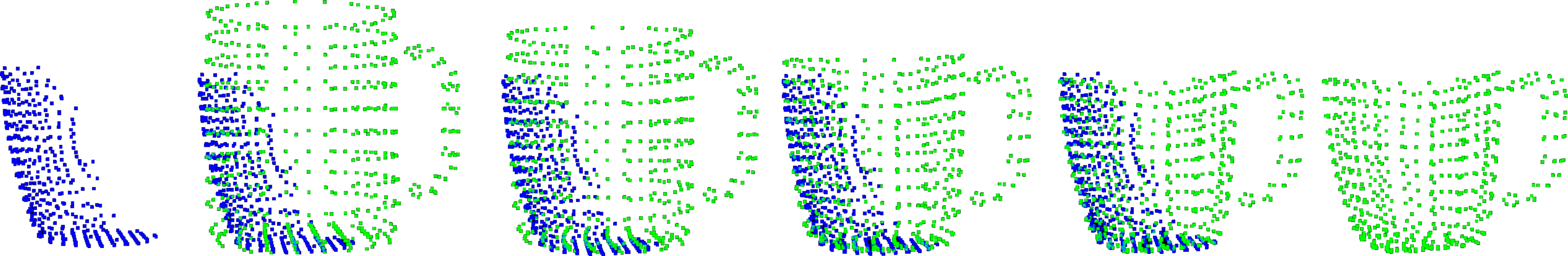}\vspace{10pt}
    \includegraphics[width=1.0\linewidth]{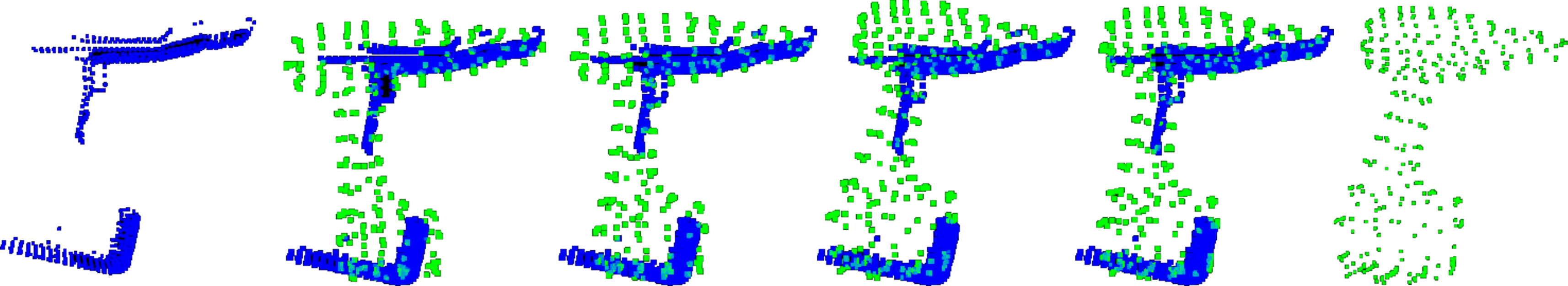}     
    \caption{\footnotesize Given a partial view of the object (leftmost), the canonical objects are deformed for the \textit{Mug} and \textit{Drill} categories. The resulting point cloud is shown rightmost.}
    \label{fig:interpolation}
\end{figure}

Fig~\ref{fig:interpolation} shows, for both categories, how the canonical shape is deformed to a single view of an observed instance. 
Note that our method is able to reconstruct occluded parts, as for example, the handles of the mugs.

For the evaluation, we take the noiseless fully-observed shape as the ground truth and the following error function:
\begin{equation}
    \label{eq:Point_Error}
    E(\mathbf{D}, \mathbf{O}^*) = \frac{1}{N}\sum_{n=1}^{N}{\min_m\left(\|\mathbf{O}^*_n-\mathbf{D}_m\|^2\right)}
\end{equation}
where $\mathbf{D}$ is the transformed canonical shape and $\mathbf{O}^*$ is the ground truth shape. 
To compute the final error, we average the errors resulting from each partial view and from each test sample.

\begin{figure}[b]
	\centering
	\includegraphics[height=2.cm]{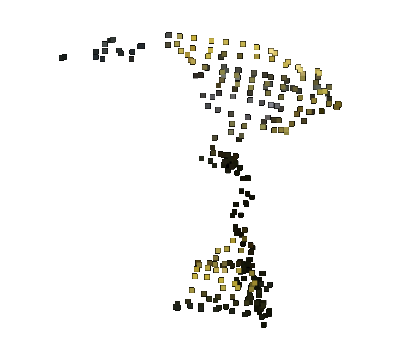}\hfill
	\includegraphics[height=2.cm]{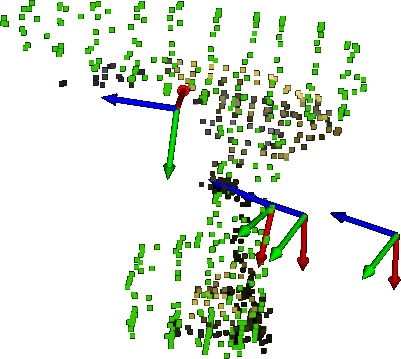}\hfill
	\includegraphics[height=2.cm]{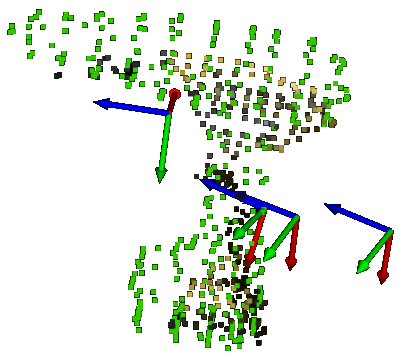}\hfill
	\includegraphics[height=2.cm]{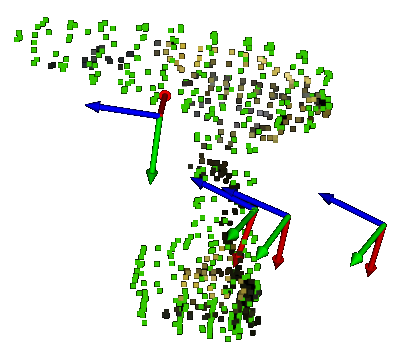}\hfill
	\includegraphics[height=2.cm]{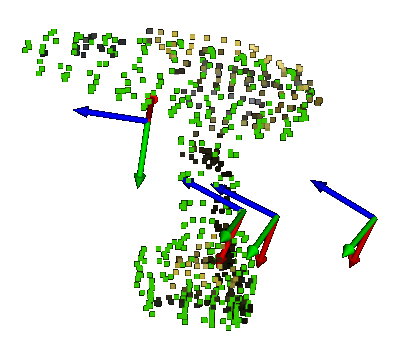}\hfill
	\includegraphics[height=2.cm]{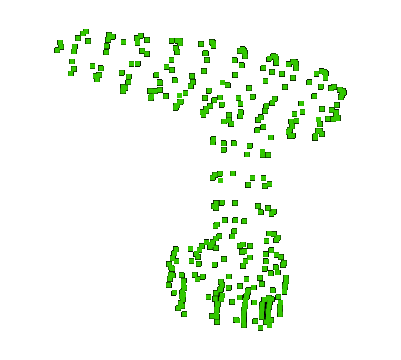}
	\caption{\footnotesize Registration of a partially occluded point cloud of an object coming from real sensory data. The input point cloud is at the top leftmost while the inferred shape is at the bottom rightmost.}
	\label{fig:eval_CLS}
\end{figure}

\begin{figure*}
	\centering
	Fully observed shapes \hspace{150pt} Partially observed shapes \\ \vspace{3pt}
	\includegraphics[width=0.26\linewidth]{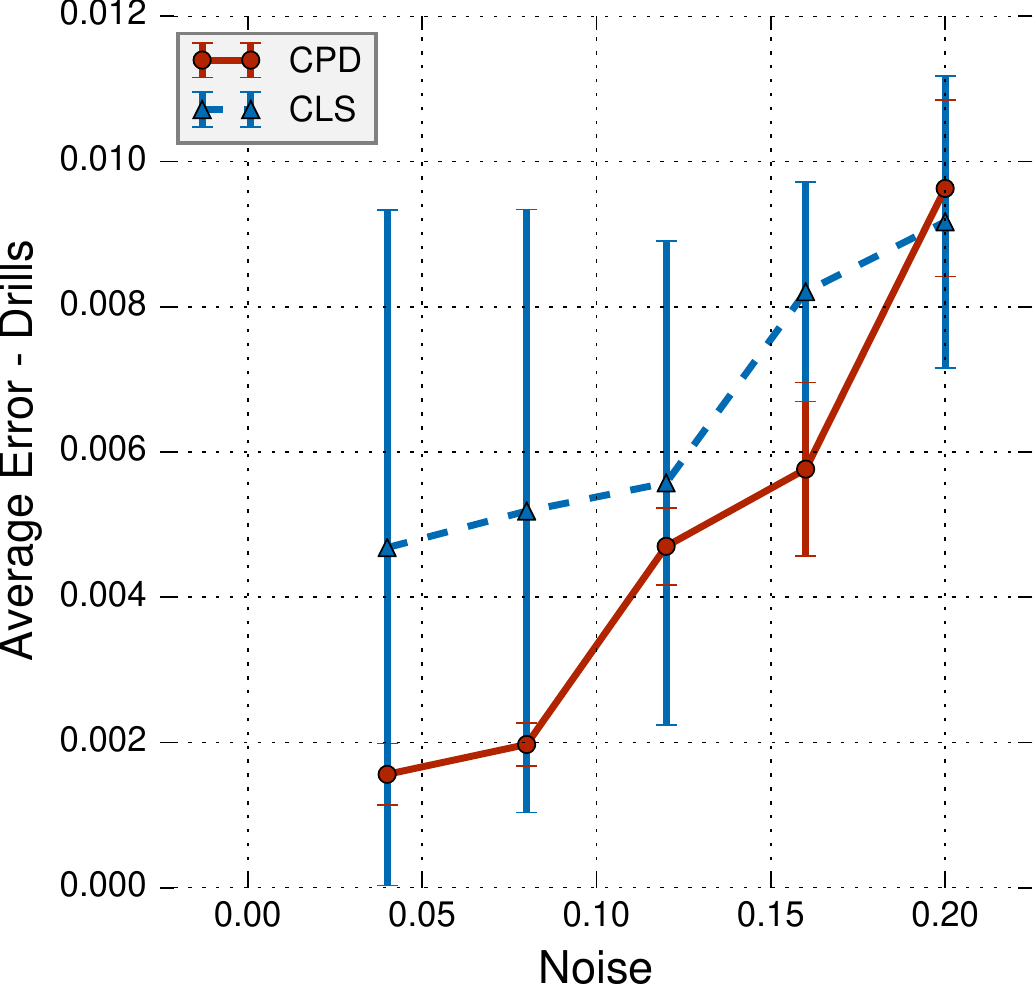}     
	\includegraphics[width=0.24\linewidth]{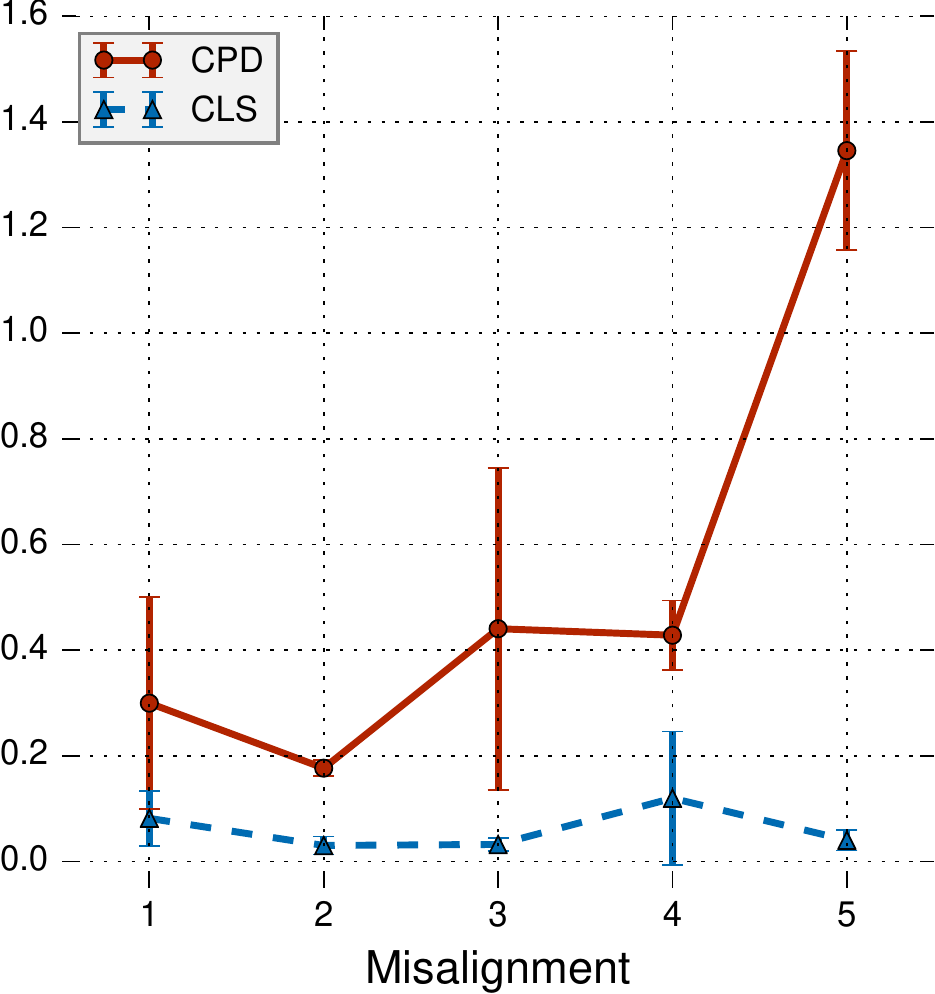} 
	\includegraphics[width=0.24\linewidth]{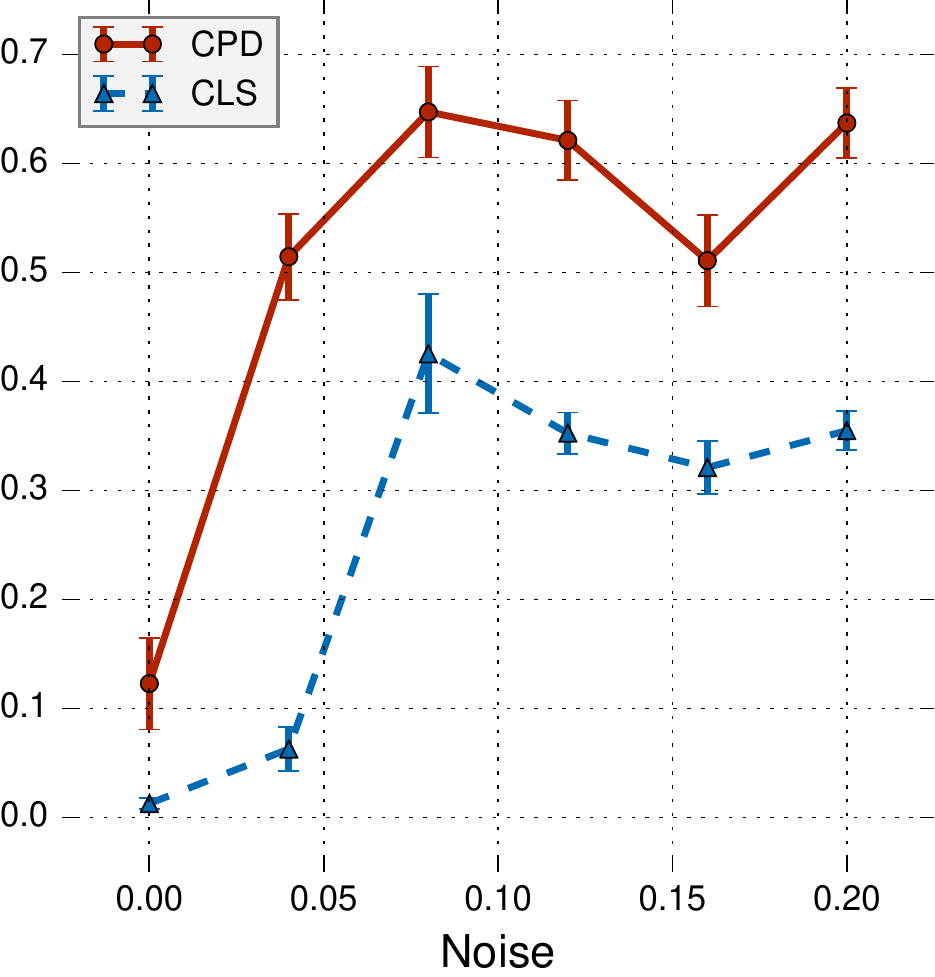} 
	\includegraphics[width=0.24\linewidth]{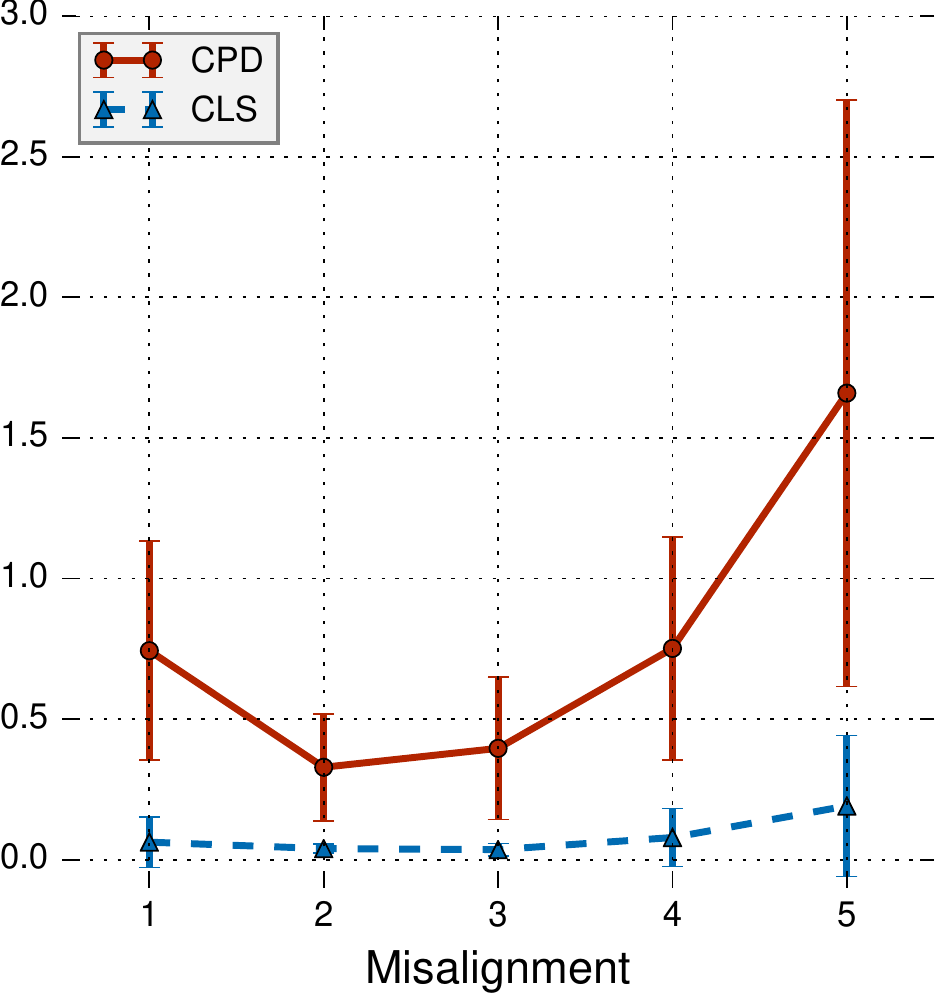}        
	\includegraphics[width=0.25\linewidth]{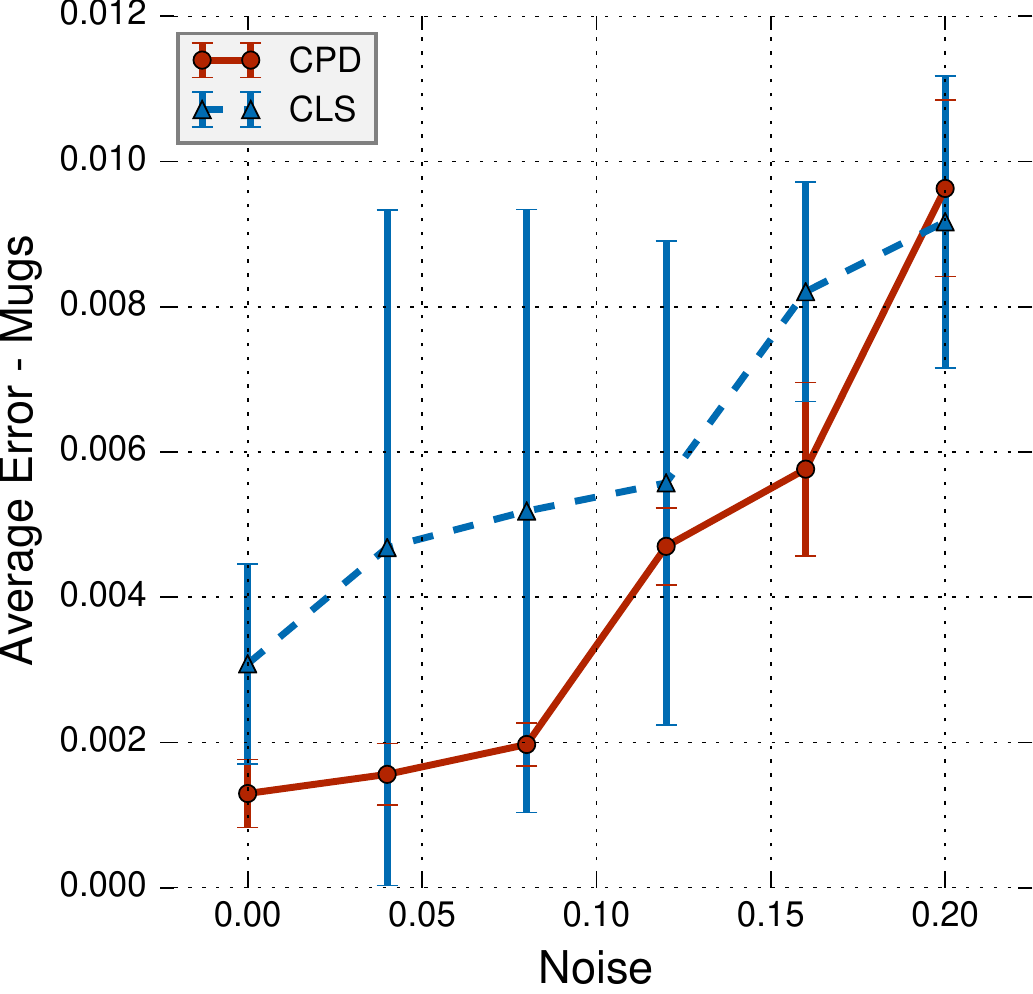}       
	\includegraphics[width=0.24\linewidth]{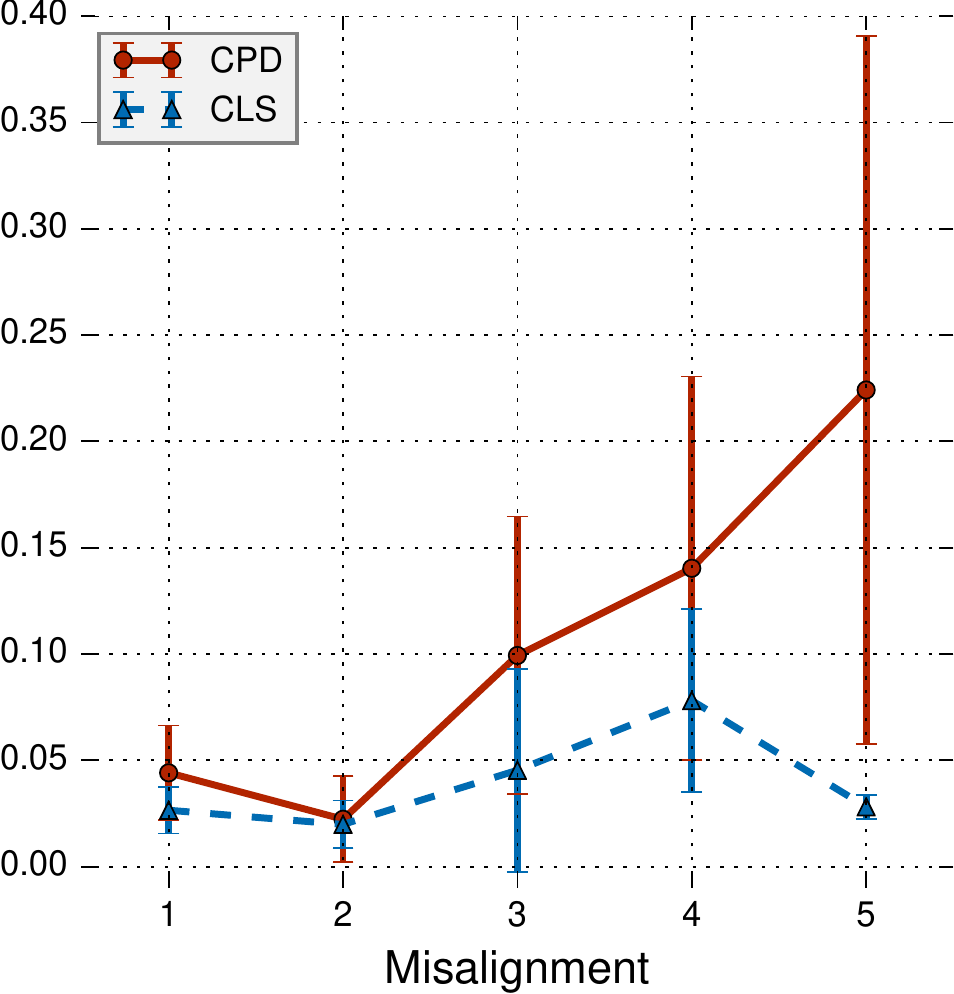} 
	\includegraphics[width=0.24\linewidth]{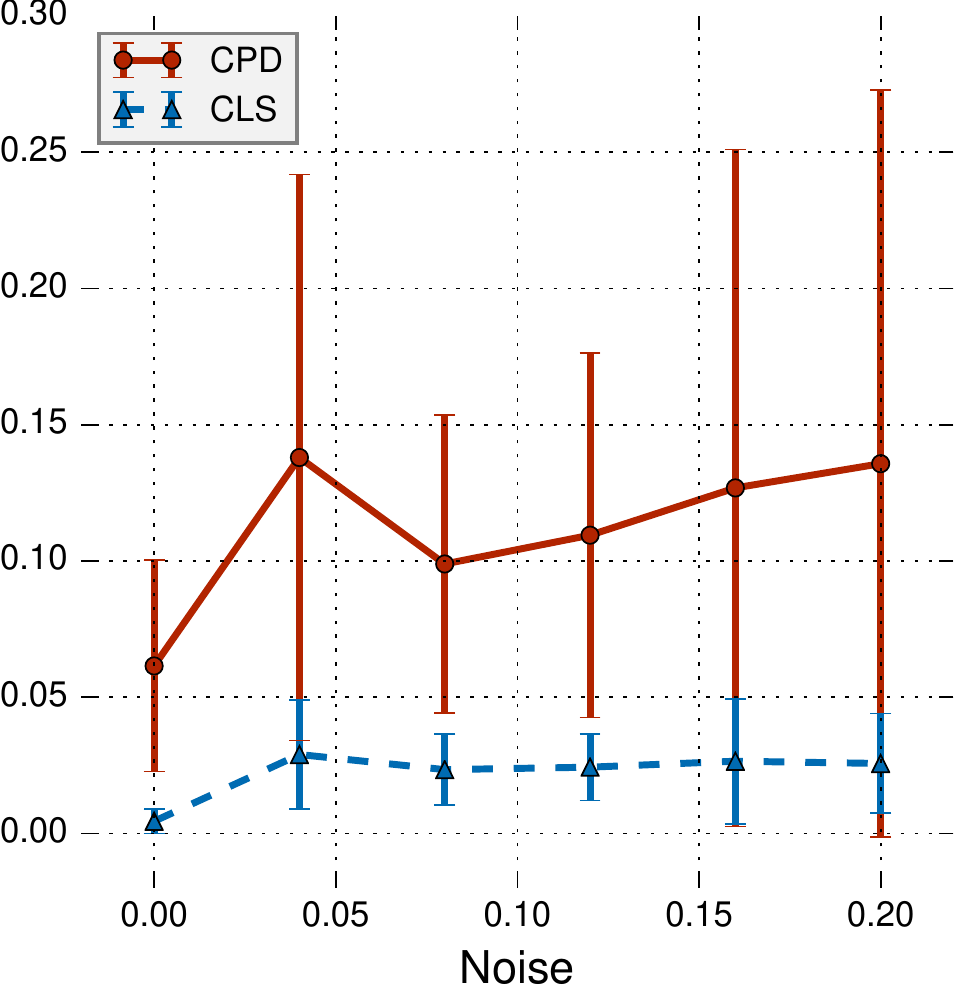} 
	\includegraphics[width=0.24\linewidth]{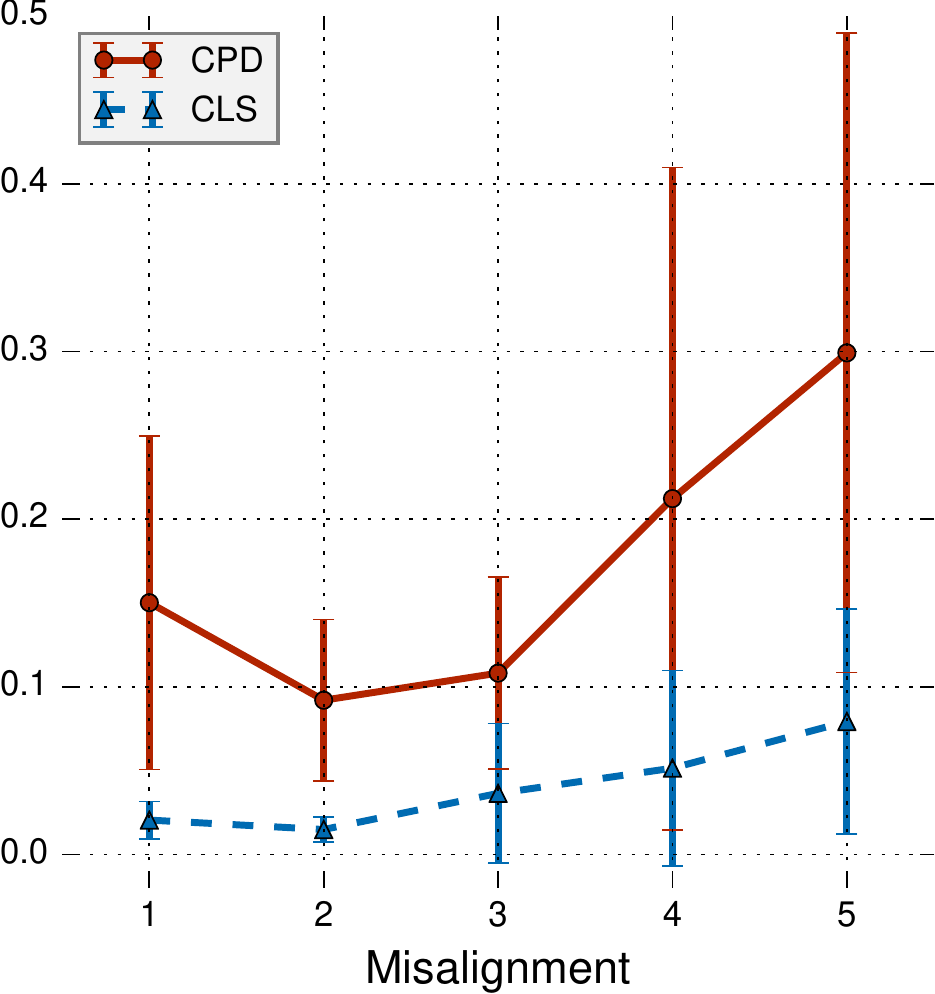} 
	\caption{\footnotesize A comparison of our method (CLS dotted blue line) against CPD (red line). The first row of plots corresponds to the \textit{drill} category while the second row refers to the \textit{cup} category. On each case, the first two plots show the results on fully observed shapes, while the last two, on partially observed shapes. Our method outperforms CPD on partial views of objects and misaligned fully observed shapes.}
	\label{fig:results_registration}
\end{figure*}

The results of our experiments are plotted in Fig. \ref{fig:results_registration}. 
We use the abbreviation CLS (Categorical Latent Space) for referring to our method.
When a shape is fully visible with or without noise, CPD outperforms our method. 
This is rather to be expected as CPD is the source of training data from which we build our registration model.
On the other hand, when the observed shape is misaligned, our method outperforms CPD, which can be explained by the additional rigid transformation component of our method.
Moreover, when the shape becomes partially occluded, the method also outperforms CPD thanks to the topological information that lies in the latent space, which is not available in CPD.

For testing the grasping transfer, we conducted a set of experiments using the five-fingered Schunk hand, which has a total of $20$ Degrees of Freedom (DoF) from which $9$ are fully actuated. 
The experiments were carried out in the Gazebo simulation environment. 
The tests were performed on the \textit{Drill} category.
Due to the reduced number of meshes, we used cross-validation and created five transformation models leaving two testing samples for each transformation model.
A trial was defined as successful if after the execution of the motion, the drill was held by the robotic hand.
We got a success rate of $0.8$ on grasping the drills.

Failure cases occurred mainly for one reason, i.e., the size of the testing object was too small compared to the canonical shape. 
This indicated that the transformation model was not general enough, mainly due to the reduced number of training samples.
The small size of the objects also caused collisions between the fingers at the moment of grasping the handle.
However, we expect to tackle this problem in the future by inferring information from the shape such as the number of fingers required for grasping.

We tested our approach with real sensory data coming from a Kinect v2 sensor~\citep{fankhauser}.
We trained a single latent space model with $9$ training samples and the canonical model.
Our method was able to generate plausible category-alike shapes and grasping poses in each of three different trials.
The inference in average took \unit[6]{s} on a Corei7 CPU \unit[2.6]{GHz} with 16GB RAM.
This demonstrates that our method can be used on-board in real robotic platforms.
The result of the registration is presented in Fig.~\ref{fig:eval_CLS}.
\section{Conclusion}
\label{sec:conclusion}
We presented a novel approach for transferring grasping skills based on a latent space non-rigid registration method, which is able to handle fully observed and partially observed shapes. 
We demonstrated its robustness especially to noise and occlusion. 
We evaluated our method on two sets of shapes and found that while the method performed slightly worse than conventional CPD on noiseless, fully observed shapes, when shapes were partially occluded, our method was much more able to recover the true underlying shape and reported lower average error.
Successful application of the non-rigid registration for transferring grasping skills was also demonstrated both with synthetic and real sensory data.

In the future, we would like to extend our approach to more complex objects that require a part-based modeling approach.

\section*{Acknowledgements}
This work was supported by 
the European Union's Horizon 2020 Programme under Grant Agreement 644839 (CENTAURO)
and the German Research Foundation (DFG) under the grant BE 2556/12 ALROMA in priority programme SPP 1527 Autonomous Learning.


\bibliographystyle{IEEEtranN}
\bibliography{icra18_grasplearn}

\end{document}